%% file: Anon_SignGAN.tex
\documentclass[a4paper, 10pt, conference]{ieeeconf}      % Use this line for a4
\usepackage[dvipsnames]{xcolor}      
\usepackage[breaklinks=true,bookmarks=false,colorlinks,citecolor=green,linkcolor=green]{hyperref}
                                                         
\usepackage{FG2021}

\FGfinalcopy % *** Uncomment this line for the final submission

\IEEEoverridecommandlockouts                              % This command is only
\usepackage{graphics} % for pdf, bitmapped graphics files
\usepackage{epsfig} % for postscript graphics files
\usepackage{mathptmx} % assumes new font selection scheme installed
\usepackage{times} % assumes new font selection scheme installed
\usepackage{amsmath} % assumes amsmath package installed
\usepackage{amssymb}  % assumes amsmath package installed
\usepackage{bm}

\usepackage{caption}
\usepackage{booktabs}

\usepackage[nolist]{acronym} % Acronyms from file
\input{acronyms} % Self-created acronyms

\def\etal{\emph{et al. }}

\newcommand{\methodname}{\textsc{AnonySign}}

\newcommand{\SE}{\textsc{SE}}
\newcommand{\PE}{\textsc{PE}}
\newcommand{\G}{\textsc{G}}
\newcommand{\DP}{\textsc{D}_\mathcal{P}}
\newcommand{\DS}{\textsc{D}_\mathcal{S}}
\newcommand{\CS}{\mathcal{C}^\mathcal{S}}

\DeclareMathAlphabet{\mathcal}{OMS}{cmsy}{m}{n}
\newcommand{\mc}{\mathcal}

\usepackage{dblfloatfix}

\usepackage{mathtools}
\DeclarePairedDelimiterX{\infdivx}[2]{(}{)}{%
  #1\;\delimsize\|\;#2%
}

\usepackage{titlesec}
\def\FGPaperID{21} % *** Enter the FG2021 Paper ID here

\title{\LARGE \bf \methodname{}: Novel Human Appearance Synthesis for \\ Sign Language Video Anonymisation}

\author{\parbox{16cm}{\centering
    {\large Ben Saunders, Necati Cihan Camgoz, Richard Bowden}\\
    {\normalsize
    University of Surrey\\}}
}
\begin{document}

\ifFGfinal
\thispagestyle{empty}
\pagestyle{empty}
\else
\author{Anonymous FG2021 submission\\ Paper ID \FGPaperID \\}
\pagestyle{plain}
\fi
% \maketitle

\twocolumn[{%
\renewcommand\twocolumn[1][]{#1}%
\maketitle
\begin{center}
    \centering
    \includegraphics[width=.89\textwidth]{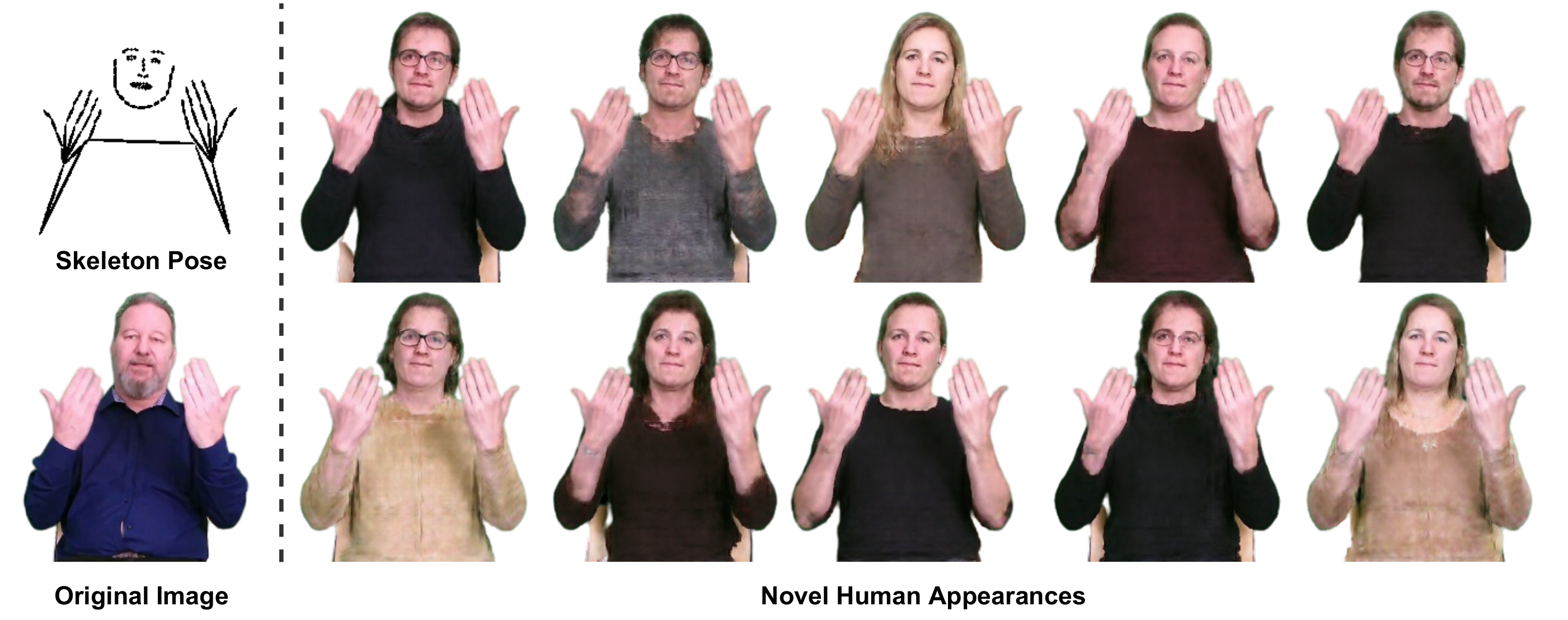}
    \captionof{figure}{Novel Appearance Human Synthesis examples generated by \methodname{}}
    \label{fig:intro}
\end{center}%
}]

\begin{abstract}

The visual anonymisation of sign language data is an essential task to address privacy concerns raised by large-scale dataset collection. Previous anonymisation techniques have either significantly affected sign comprehension or required manual, labour-intensive work.

In this paper, we formally introduce the task of \ac{slva} as an automatic method to anonymise the visual appearance of a sign language video whilst retaining the meaning of the original sign language sequence. To tackle \ac{slva}, we propose \methodname{}, a novel automatic approach for visual anonymisation of sign language data. We first extract pose information from the source video to remove the original signer appearance. We next generate a photo-realistic sign language video of a novel appearance from the pose sequence, using image-to-image translation methods in a conditional variational autoencoder framework. An approximate posterior style distribution is learnt, which can be sampled from to synthesise novel human appearances. In addition, we propose a novel \textit{style loss} that ensures style consistency in the anonymised sign language videos.

We evaluate \methodname{} for the \ac{slva} task with extensive quantitative and qualitative experiments highlighting both realism and anonymity of our novel human appearance synthesis. In addition, we formalise an anonymity perceptual study as an evaluation criteria for the \ac{slva} task and showcase that video anonymisation using \methodname{} retains the original sign language content.

\end{abstract}

%%%%%%%%%%%%%%%%%%%%%%%%%%%%%%%%%%%%%%%%%%%%%%%%%%%%%%%%%%%%%%%%%%%%%%%%%%%%%%%%

\section{Introduction} \label{sec:introduction}

% Sign Language intro
Sign languages are rich visual languages, and the main mode of communication for Deaf communities. As with the computational study of spoken languages, sign language research requires large-scale datasets to effectively tackle the tasks of \ac{slt} and \ac{slp}. However, due to privacy concerns regarding the visual nature of sign language datasets, there is a hesitancy to contribute to data collection \cite{bragg2020exploring}.

% Sign Language Video Anonymisation task

In this work, we formally introduce the task of \acf{slva} as an automatic method to anonymise the visual appearance of a sign language video whilst retaining the original sign language content. Full anonymisation requires two aspects of the signer to be unidentifiable: visual appearance and signing style. In this work, we focus on the former and leave the latter task of signing style anonymisation for future work.

Visual anonymisation of sign language datasets has been shown to increase the willingness of data participation due to the removal of signer-specific appearance \cite{bragg2020exploring}. However, previous approaches to sign language video anonymisation, such as pixelation \cite{rudge2018analysing}, blackening \cite{bleicken2016using} and greyscale filtering \cite{bragg2020exploring}, have fallen woefully short of the mark. These methods significantly affect sign comprehension or worse, in the case of animoji filtering \cite{bragg2020exploring}, deride the signers appearance. 

% In addition, anonymisation can enable online privacy for sign language users. 

% Propose A-SignGAN
To tackle \ac{slva}, we propose \methodname{}, a novel automatic method that achieves visual anonymisation of sign language data whilst retaining sign language comprehension. We build \methodname{} as a conditional variational generative adversarial network (cVAE-GAN) \cite{bao2017cvae} with a combination of \acp{vae} and \acp{gan}. To remove the original signers appearance, we first extract a skeleton pose sequence from the source sign language video via pose estimation \cite{cao2018openpose}. 

% VAE sampling
During training, we use a style encoder to encode style images into a style code that represents signer-specific appearance features such as facial attributes and body shape. Implicitly, \methodname{} learns an approximate posterior style distribution from the space of all pose-independent style images, in the framework of a conditional \ac{vae}. We sample from this distribution at inference time to generate multiple novel human appearances not seen during training. To ensure both style consistency and the disentanglement of pose and style, we propose a novel \textit{style loss} that minimises the distance between style distributions extracted from the style image and generated image, respectively.

% Photo-Realistic generation
Finally, we reconstruct a photo-realistic human appearance in the given pose by using a generator with multiple \ac{adain} layers \cite{huang2017arbitrary}. \ac{adain} learns the affine transform parameters relating to the desired style of the sampled style code. This enables \methodname{} to puppeteer the novel signer appearance with the original sign language pose sequence, resulting in a visual anonymisation of the original video that retains all sign language content.

% Evaluation
We evaluate \methodname{} for the \ac{slva} task using a large-scale isolated sign language dataset \cite{ebling2018smile}. We conduct extensive quantitative and qualitative evaluation that showcase the effectiveness of \methodname{} for generating novel signer appearances. In addition, we formalise an anonymity perceptual study as an evaluation criteria for the \ac{slva} task, drawing parallels to super-recogniser tests \cite{dunn2020unsw}. Finally, we show that our \ac{slva} approach retains the original sign language content using a downstream \ac{slt} task.

The \textbf{contributions} of this work can be summarised as:

\begin{itemize}
    \item The formal introduction of the automatic \acf{slva} task
    \item \methodname{}, a novel automatic \ac{slva} method that uses a conditional variational autoencoder framework to generate novel human appearances not seen during training
    % \item \methodname{}, the first pose-conditioned human synthesis model to generate novel human appearances not seen during training \todo{Which of these two sounds better?}
    \item A novel style loss that enhances style consistency in generated sign language videos
\end{itemize}

The rest of this paper is organised as follows: In Section~\ref{sec:related}, we review the literature in sign language anonymisation and human synthesis. In Section~\ref{sec:methodology}, we outline the proposed \methodname{} approach. Section~\ref{sec:experiments} presents quantitative and qualitative evaluation, whilst Section \ref{sec:conc} concludes.

\section{Related Work}
\label{sec:related}

\subsection{\textbf{Sign Language Production}}

Combining the domains of vision and language, computational sign language research has been prominent for the last 30 years \cite{tamura1988recognition}. Previous research has focused on isolated sign recognition \cite{ozdemir2016isolated}, \ac{cslr} \cite{camgoz2017subunets} and \ac{slt} \cite{camgoz2018neural,camgoz2020sign}. \acf{slp}, the translation from spoken language to sign language, has been traditionally tackled using graphical avatars \cite{mcdonald2016automated,ebling2015bridging} with a recent growth of deep learning-based approaches \cite{saunders2020adversarial,saunders2020progressive,saunders2021continuous,stoll2020text2sign,zelinka2020neural}. 

Photo-realistic \ac{slp} has been proposed to overcome the difficulty of skeleton pose comprehension \cite{stoll2020text2sign,cui2019deep}. Saunders \etal proposed \textsc{SignGAN} \cite{saunders2020everybody} to generate photo-realistic continuous sign video sequences, overcoming the previous lack of detail in hand synthesis. In this work, we overcome the privacy issues of signer pupeteering with the synthesis of novel human appearances. In addition to anonymisation, the ability to sample multiple novel appearances further enables diversity in signer generation, which has been highlighted as important by Deaf focus groups \cite{kipp2011assessing}.

\subsection{\textbf{Sign Language Video Anonymisation}}

The field of \ac{nmt} has seen incredible recent progress with the availability of large-scale datasets \cite{camgoz2021content4all,chelba2014one,marcus1993building}. However, there has been a lack of large-scale computational sign language datasets collected. Due to its visual medium, the collection of large-scale sign language corpora requires the storing of easily identifiable video data. Bragg \etal suggests this introduces privacy concerns over data misuse, that significantly impacts the collection of sign language datasets \cite{bragg2020exploring}.

There has been little previous research into sign language video anonymisation. Video effects of blackening \cite{bleicken2016using} and pixelation \cite{rudge2018analysing} have been used to conceal sensitive information, which is unfeasible for extension to full data corpora. Bragg \etal \cite{bragg2020exploring} suggest the use of greyscale or animoji filtering for video anonymisation, but neither provides full signer anonymity and both significantly impact sign comprehension. Focus groups with Deaf participants have suggested the use of actors or avatars to reproduce the data \cite{singleton2014toward}, but this requires labour-intensive work \cite{isard2020approaches} and often results in non-realistic production \cite{kipp2011assessing}. 

Previously, the idea of complete sign language video anonymisation has been deemed impossible \cite{isard2020approaches,quer2019handling}, due to the required visibility of both face and hands. Disputing this claim, we formalise the task of \acf{slva} to promote future research and propose a novel automatic visual anonymisation method.

\begin{figure*}[h]
    \centering
    \includegraphics[width=1.0\linewidth]{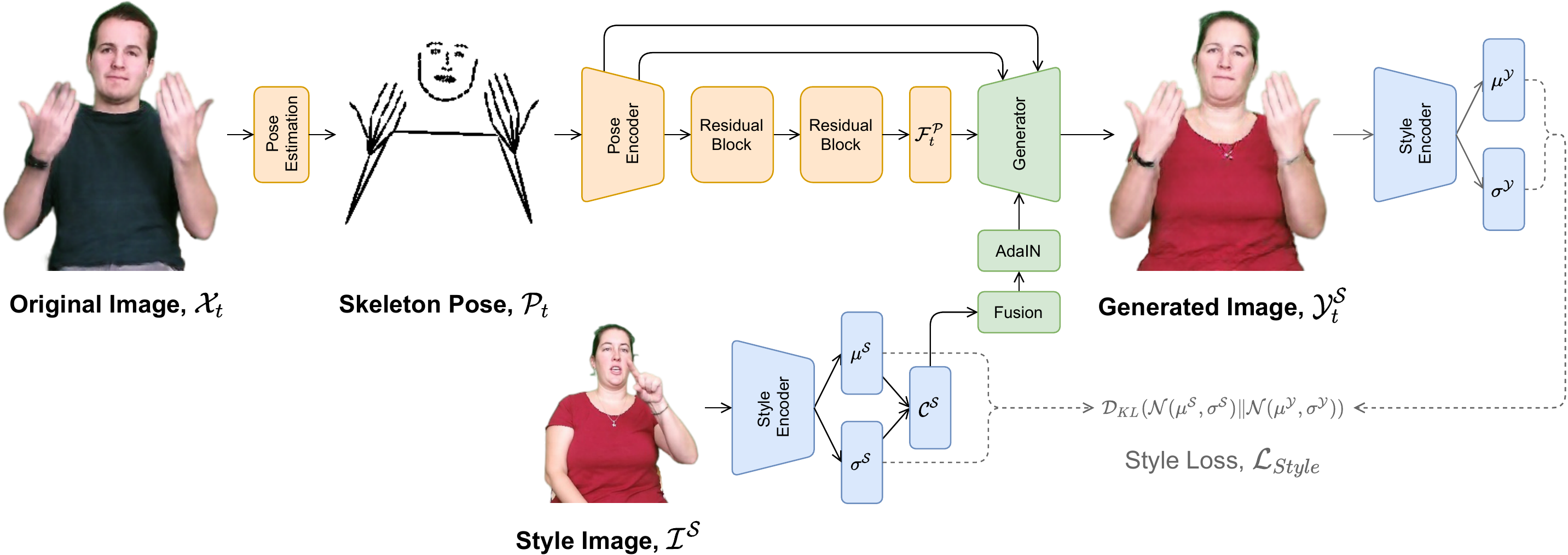}
    \caption{An overview of the \methodname{} network for Sign Language Video Anonymisation (SLVA). At training, skeleton pose, $\mc{P}_{t}$, is extracted from an original image, $\mc{X}_{t}$ using a pre-trained \textit{Pose Estimator}. The skeleton pose is then encoded by a \textit{Pose Encoder} to a set of pose features, $\mc{F}_{\mc{P}}$. A Style Image, $\mc{I}^{\mc{S}}$, is encoded by a \textit{Style Encoder} to a style distribution, $\mc{N}(\mu_{\mc{S}},\sigma_{\mc{S}})$, where a \textit{Style Code}, $\CS{}$, is sampled using the re-parameterisation trick. A photo-realistic signer image, $\mc{Y}^{\mc{S}}$, of pose $\mc{P}$ and style $\mc{S}$ is generated by a \textit{Generator} with multiple \ac{adain} layers to embed the style from $\CS{}$. To disentangle style and pose, out \textit{Style Loss}, $\mc{L}_{Style}$, applies a KL divergence between the encoded style distributions of $\mc{I}^{\mc{S}}$, $\mc{N}(\mu^{\mc{S}},\sigma^{\mc{S}})$, and $\mc{Y}^{\mc{S}}$, $\mc{N}(\mu^{\mc{Y}},\sigma^{\mc{Y}})$. }
    \label{fig:model_overview}
\end{figure*}%

\subsection{\textbf{Pose Conditioned Human Synthesis}}

\acfp{gan} \cite{goodfellow2014generative} have achieved impressive results in pose-conditioned human synthesis, for the concurrent tasks of pose transfer \cite{chan2019everybody} and controllable body generation \cite{men2020controllable,sarkar2021humangan}. Latent space encoding has provided disentanglement of shape and appearance \cite{men2020controllable}, and body part sampling \cite{esser2018variational,sarkar2021humangan}. Concurrently, \acf{adain} \cite{huang2017arbitrary} has been used to decode style into a generated image, for controllable image synthesis \cite{karras2019style} and multimodal human synthesis \cite{men2020controllable}. We expand pose conditioned human synthesis to the generation of a novel signer appearance not seen during training.

\subsection{\textbf{Variational Autoencoders}}

\acfp{vae} are generative models that enable a sampling from the prior at inference \cite{kingma2013auto}. They have been extended to the conditional framework \cite{sohn2015learning} and combined with \acp{gan} \cite{bao2017cvae} to achieve high-quality multimodal synthesis of human pose and appearance \cite{esser2018variational,sarkar2021humangan}. Our method builds on conditional \acp{vae} to synthesise a signer with a novel appearance via latent code sampling.

\section{Methodology} \label{sec:methodology}
Given a source sign language video, $\mc{X} = (\mc{X}_{1},...,\mc{X}_{\mc{T}})$ with $\mc{T}$ frames, our goal is to generate an anonymised target video with the same sign language content, $\mc{Y} = (\mc{Y}_{1}^{\mc{S}},...,\mc{Y}_{\mc{T}}^{\mc{S}})$, where $\mc{S}$ represents a novel appearance not seen in the training data. We approach this problem as an image-to-image translation task with a conditional variational autoencoder framework. Our proposed \methodname{} architecture is outlined in Fig. \ref{fig:model_overview}.

During training, we first extract pose, $\mc{P}_{t}$ from the $t^{th}$ frame using a pose estimation model and embed it in a high-dimensional feature space, $\mc{F}_{t}^{\mc{P}}$, using a \textit{Pose Encoder}. We next learn an approximate posterior style distribution from a given style image, $\mc{I}^{\mc{S}}$, by using a \textit{Style Encoder} to encode a latent style code, $\CS{}$. Finally, a photo-realistic reconstruction of the source pose in the given style is generated, $\mc{Y}_{t}^{\mc{S}}$, using a generator with \ac{adain} layers \cite{huang2017arbitrary} that embed the style code.

To enable sampling of a novel anonymous style at inference time, we build \methodname{} using a conditional \acf{vae} framework \cite{kingma2013auto,sohn2015learning}. As formulated in latent vector models, we assume that the style of the generated image is solely dependent on the latent variable, $\CS{}$. The goal is then to learn an approximate posterior distribution, $q(\CS{} \mid \mc{I}^{\mc{S}})$, from the training style images. A style code, $\CS{}$, can be sampled from this learnt distribution at inference to synthesise a novel signer appearance. A key assumption of our model is that the latent variable $\CS{}$ depends \textit{only} on the signer style and is independent of pose. We enforce this condition by proposing a novel \textit{Style Loss} that prompts a similarity between the style distributions encoded from the style image and the generated image, which contain different poses.

\subsection{\textbf{Pose Encoder (PE)}}

To remove the original signers appearance, we first use a pre-trained pose estimation model \cite{cao2018openpose} that extracts skeleton pose, $\mc{P}_{t}$, from each source frame, $\mc{X}_{t}$, as seen on the left of Fig. \ref{fig:model_overview}. We represent both manual and non-manual features\footnote{Manual features are the hand shape and motion, whereas non-manuals are the mouthings, facial expressions, gait etc.} of the signer as joint parameters of the upper body, hands and facial landmarks. Alternatively, \methodname{} can be used as the photo-realistic output of a \ac{slp} model that translates a spoken language sentence (text) into a sign pose sequence, as in \cite{saunders2020everybody}.

We next encode the extracted pose, $\mc{P}_{t}$, into a higher-dimensional feature space, $\mc{F}_{t}^{\mc{P}}$, using a \textit{Pose Encoder}, $\PE{}(\cdot)$:
\begin{equation} \label{eq:pose_encoder}
    \mc{F}_{t}^{\mc{P}} = \PE{}(\mc{P}_{t}),
\end{equation}
where $\mc{F}_{\mc{P}}$ is the extracted pose features.

\begin{figure*}[h]
    \centering
    \includegraphics[width=0.9\linewidth]{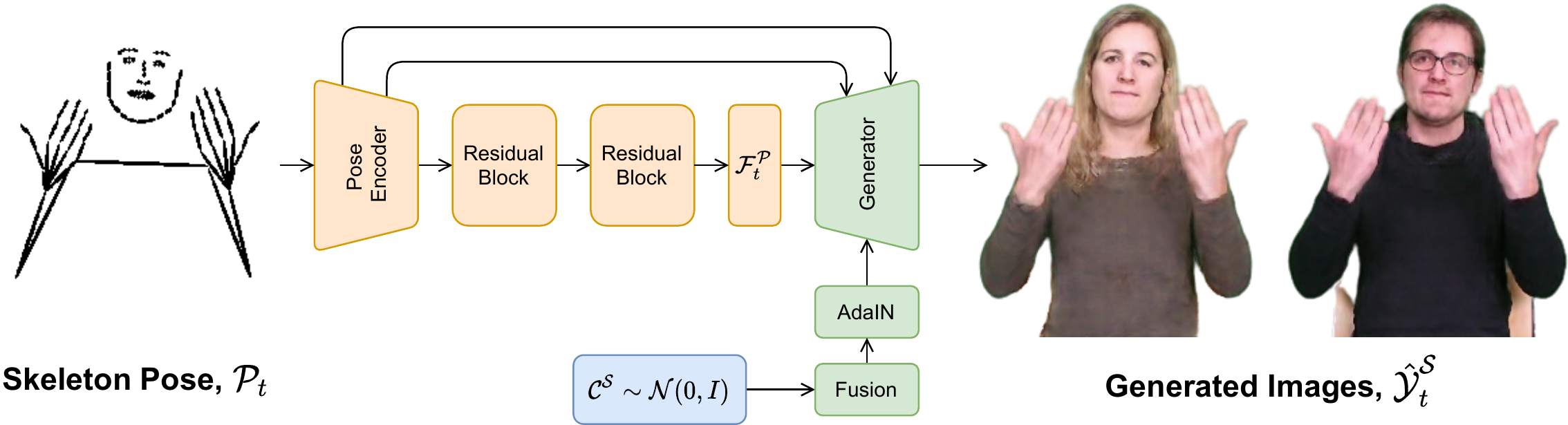}
    \caption{Novel human appearances, $\hat{\mc{Y}}^{\mc{S}}_{t}$, can be generated from \methodname{} by sampling a style code from the prior distribution, $\CS{} \sim \mc{N}(0,I)$. Varying $\mc{P}_{t}$ whilst keeping $\CS{}$ consistent enables the anonymisation of a full sign language video.}
    \label{fig:model_sampling}
\end{figure*}%

\subsection{\textbf{Style Encoder (SE)}}

During training, we extract style information from a given style image, $\mc{I}^{\mc{S}}$, using a \textit{Style Encoder}, $\SE{}(\cdot)$, as seen at the bottom of Fig. \ref{fig:model_overview}. We represent the appearance of the signer as a latent style code, $\CS{}$, of dimension $\mc{D}_{\mc{C}}$. We define style as the person-specific appearance aspects such as body shape, face, hair or clothing. The pose information is signer-invariant and is thus learnt independently from style. During training, we randomly select $\mc{I}^{\mc{S}}$ from the training data that contains identical signer style but different pose to the source frame, $\mc{X}^{\mc{S}}_{t}$.

As for a \ac{vae} \cite{kingma2013auto}, the style encoder learns an approximate posterior style distribution, $P(\CS{} \mid \mc{I}^{\mc{S}})$, from the space of pose-independent style images, $\mc{I}^{\mc{S}}$. We assume the style code distribution to be Gaussian, $p(\CS{} \mid \mc{I}^{\mc{S}}) \equiv \mc{N}(\mu,\sigma)$ and thus represent style with parameters $\mu_{\mc{S}},\sigma_{\mc{S}} \in \mathbb{R}^{\mc{D}_{\mc{C}}}$. Using the re-parameterisation trick, we can sample the style code, $\CS{}$, as:
\begin{equation} \label{eq:style_encoder}
    \CS{} \sim \mc{N}(\mu^{\mc{S}},\sigma^{\mc{S}}) = \SE{}(\mc{I}^{\mc{S}})
\end{equation}
\textbf{Style Loss:} To ensure the style code is disentangled from pose information, we propose a style consistency loss. As seen on the right of Fig. \ref{fig:model_overview}, a KL divergence is measured between the style distribution encoded from the style image, $\mc{N}(\mu^{\mc{S}},\sigma^{\mc{S}})$, and the generated image, $\mc{N}(\mu^{\mc{Y}},\sigma^{\mc{Y}})$, as:
\begin{equation} \label{eq:style_loss}
    \mc{L}_{Style} = \mc{D}_{KL}\infdivx{\mc{N}(\mu^{\mc{S}},\sigma^{\mc{S}})}{\mc{N}(\mu^{\mc{Y}},\sigma^{\mc{Y}})},
\end{equation}
where $\mc{D}_{KL}\infdivx{p}{q}$ is the Kullback-Leibler divergence \cite{kullback1951information} between the probability distributions $p(x)$ and $q(x)$. This provides extra supervision to $\SE{}$ whilst also enforcing style consistency in the generated image.

\subsection{\textbf{Photo-Realistic Image Generator (G)}}

Given the encoded pose features, $\mc{F}_{\mc{P}}$, and style code, $\CS{}$, a generator model, $\G{}(\cdot)$, decodes a photo-realistic image, $\mc{Y}_{t}^{\mc{S}}$, with the corresponding pose  and style:
\begin{equation} \label{eq:generator}
    \mc{Y}_{t}^{\mc{S}} = \G{}(\mc{F}_{t}^{\mc{P}},\CS{})
\end{equation} 
To embed the style from $\CS{}$ into the generated image, we use multiple \acf{adain} layers~\cite{huang2017arbitrary}. \ac{adain} layers first apply an instance normalisation \cite{ulyanov2017improved}, followed by a further learnt feature style normalisation. This encourages the features to have similar statistics to the desired style using the learnt affine transform parameters scale, $\mu_{\mc{A}}$, and shift, $\sigma_{\mc{A}}$, as:
\begin{equation}
    \textsc{AdaIN}(x) = \sigma_{\mc{A}} \cdot \frac{x - \mu(x)}{\sigma(x)} + \mu_{\mc{A}},
\end{equation}
where $\mu(x)$ and $\sigma(x)$ are calculated from the pre-normalised feature channel. We apply an additional fusion module to extract the affine parameters from the style code \cite{men2020controllable}. Additionally, we utilise multiple skip connections from the encoded pose layers, to enable a direct pose conditioning.

\subsection{\textbf{Training Details}}

We use the re-parameterisation trick \cite{kingma2013auto} to enable a differentiable pipeline, jointly training \PE{}, \SE{} and \G{}. Expanding from the standard image-to-image framework \cite{isola2017image,wang2018high}, we train \G{} with the objective $\mc{L}_{Total}$, a combination of adversarial, perceptual and prior losses balanced by their respective weights, $\lambda_{*}$.

\textbf{Adversarial Loss:} To enable the generation of realistic images of both pose and style, we apply a multi-scale adversarial loss \cite{wang2018high}. Following \cite{zhu2019progressive}, we use two discriminators: A pose discriminator, $\DP{}$, that prompts the pose alignment of the generated image, $\mc{Y}_{t}^{\mc{S}}$, with the target pose, $\mc{P}_{t}$; and a style discriminator, $\DS{}$, that ensures a style similarity between the generated image, $\mc{Y}_{t}^{\mc{S}}$ and the style image, $\mc{I}^{\mc{S}}$. The full adversarial loss is formulated as $\mc{L}_{Adv} = \lambda_{DP} \mc{L}_{DP}(\G{},\DP{}) + \lambda_{DS} \mc{L}_{DS}(\G{},\DS{})$.

\textbf{Hand Keypoint Loss:} To ensure high-quality hand synthesis for sign language comprehension, we apply the hand keypoint loss proposed in \cite{saunders2020everybody}. A hand keypoint discriminator, $\textsc{D}_\mathcal{H}$, is applied that operates in the keypoint space to overcome motion blur. Formally, $\mc{L}_{Hand} = \lambda_{DH} \mc{L}_{DH}(\G{},\textsc{D}_\mathcal{H})$.

\textbf{Perceptual Loss:} We include two feature-matching perceptual losses: $\mc{L}_{FM}(\G{},\DP{})$, the discriminator feature-matching loss presented in pix2pixHD \cite{wang2018high}; and $\mc{L}_{VGG}(\G{},\DP{})$, the perceptual reconstruction loss that compares pretrained VGGNet \cite{simonyan2014very} features at multiple layers of the network. The full perceptual loss is formulated as $L_{Perc} = \lambda_{FM} \mc{L}_{FM}(\G{},\DP{}) + \lambda_{VGG} \mc{L}_{VGG}(\G{},\DP{})$.

\textbf{Prior Loss:} To ease the sampling at inference, we encourage the style encoding, $\SE{}(\mc{I}^{\mc{S}})$ to be close to a standard Gaussian distribution, where the prior distribution on the style code, $\CS{}$, is assumed to be $\mc{N}(0,I)$. The prior loss is formulated as $\mc{L}_{Prior} = \lambda_{KL} \mc{D}_{KL} \infdivx{\SE{}(\mc{I}^{\mc{S}})}{\mc{N}(0,I)}$.

Our full objective, $\mc{L}_{Total}$, is a weighted sum of these losses, alongside our proposed style loss (Eq. \ref{eq:style_loss}):
\begin{align} \label{eq:loss_total}
    \mc{L}_{Total} = \mc{L}_{Adv} + \mc{L}_{Hand} + \mc{L}_{Perc} + \mc{L}_{Prior} + \lambda_{Style} \mc{L}_{Style}
\end{align}

With the re-parameterisation trick, we train \methodname{} end-to-end by optimising the parameters of \PE{}, \SE{} and \G{}. The final objective, $\mc{L}_{Total}$, is minimised with respect to \PE{}, \SE{} and \G{}, whilst maximised against $\DP{}$, $\DS{}$ and $\textsc{D}_\mathcal{H}$. 

% For speed, we pre-compute OpenPose features on the training images and read them directly as input.

\subsection{\textbf{Inference: Sampling Novel Appearances}}

During inference, we \textit{sample} a style code from the prior distribution, $\CS{} \sim \mc{N}(0,I)$, to synthesise a novel signer appearance in a given pose, as seen in Fig. \ref{fig:model_sampling}. A style can also be extracted from the approximate posterior distribution, as in Fig. \ref{fig:model_overview}, by encoding a given style image, $I^{\mc{S}}$, using the style encoder, \textit{i.e.}, $\CS{} = \mu_\mc{S}$, where $\mu_{\mc{S}}, \sigma_{\mc{S}} = SE(\mc{I}^{\mc{S}})$. A novel human appearance can be synthesised in this sampled style, $\hat{\mc{Y}}^{\mc{S}}_{t}$. By keeping $\CS{}$ fixed and varying the pose sequence, $\mc{P}_{1:\mc{T}}$, a full-length anonymised sign language video sequence can be generated that retains the sign language content of the original video, $\hat{\mc{Y}}^{\mc{S}}_{1:\mc{T}}$.

% \todo{Check if this could work for a new person? So single style image for new person - NOPE}

\section{Experiments} \label{sec:experiments}

In this section, we evaluate the performance of \methodname{} for the \ac{slva} task. We first outline our experimental setup then perform quantitative, perceptual and qualitative evaluation. 

\subsection{\textbf{Experimental Setup}}

\textbf{Dataset:} We train our anonymous human synthesis model using the SMILE isolated sign language dataset \cite{ebling2018smile}. As outlined in previous work, we use a heat-map representation as pose condition and perform background segmentation \cite{saunders2020everybody}. We also add a random horizontal shifting of up to 20 pixels to provide further data augmentation.

\textbf{Implementation Details:} We train \methodname{} for the output resolution of $256 \times 256$. $\PE{}()$ is compromised of four down-sampling and six residual blocks. $\SE{}()$ contains six down-sampling convolutional layers with a latent style code dimension, $\mc{D}_{\mc{C}}$, of 64. We omit normalisation layers from $\SE{}()$ to retain the original feature mean and variance that contains important style information \cite{huang2018multimodal}. $\G{}()$ consists of four up-sampling blocks, each with a skip connection from the features of $\PE{}()$ and a respective \ac{adain} layer \cite{huang2017arbitrary}. The loss weights in Eq. \ref{eq:loss_total} are set empirically to $\lambda_{DP}=1.0$, $\lambda_{DS}=1.0$, $\lambda_{DH}=1.0$ $\lambda_{FM}=1.0$, $\lambda_{VGG}=1.0$, $\lambda_{KL}=0.01$, $\lambda_{Style}=0.01$. All parts of our network are initialized using Xavier method and trained with an Adam optimizer using default parameters and a learning rate of $2 \times 10^{-3}$.

\begin{figure*}[t!]
    \centering
    \includegraphics[width=0.99\linewidth]{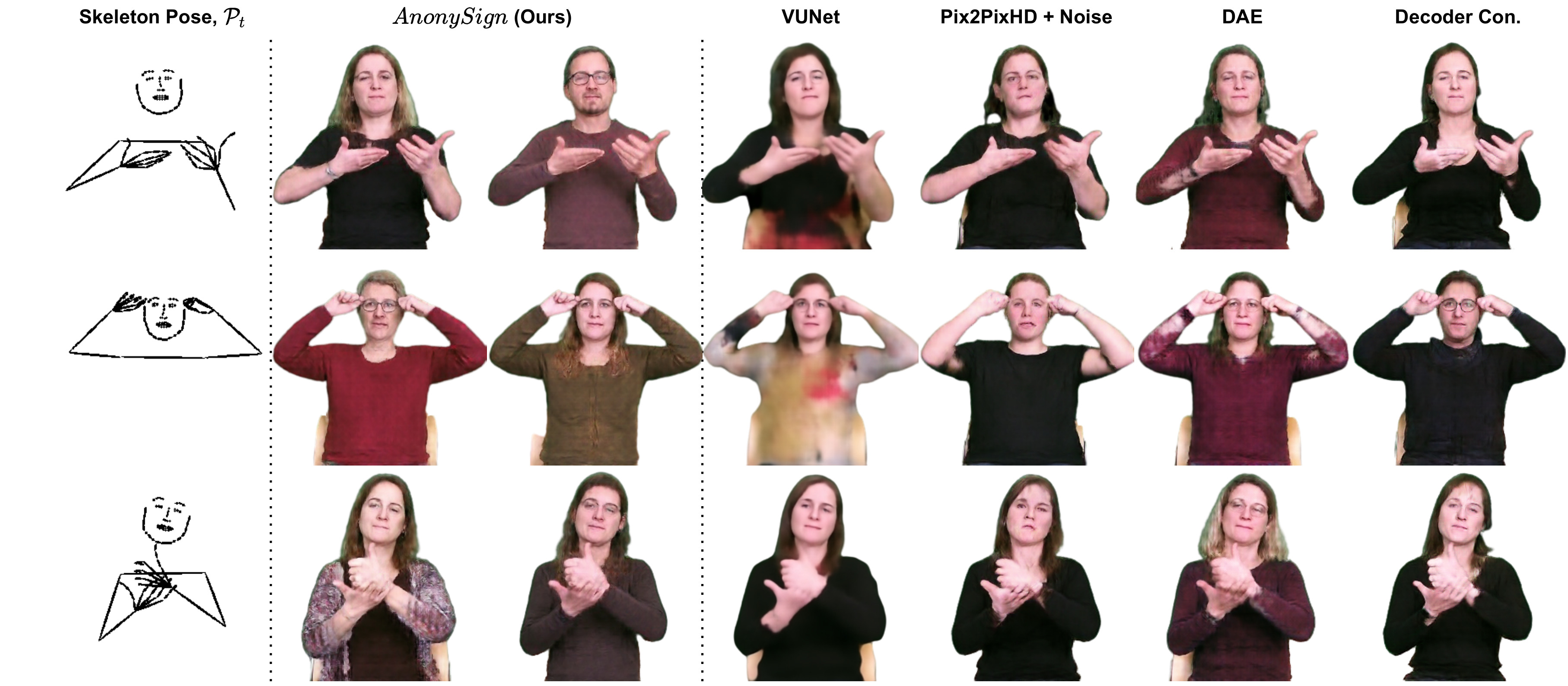}
    \caption{Qualitative baseline comparisons, highlighting the diversity and realism of our novel human appearance synthesis.}
    \label{fig:baselines}
\end{figure*}%

\textbf{Baselines:} We compare \methodname{} to the following baseline methods: 1) \textbf{VUNet} \cite{esser2018variational} performs disentanglement between shape and appearance and provides conditional human synthesis. Appearance can be sampled from the stochastic latent representation with a shape preservation. 2) \textbf{Pix2PixHD + Noise} \cite{wang2018high} is a stochastic extension to pix2pixHD, with random noise added to the input alongside pose. We tile sampled noise, $z \in \mathbb{R}^{\mc{D}_{\mc{C}}}$, across the $\mc{D}_{\mc{P}}$ pose condition to create a $\mc{D}_{\mc{P}} + \mc{D}_{\mc{C}}$ channel input vector and train pix2pixHD for a stochastic human synthesis. We note a similar baseline was used in \cite{sarkar2021humangan}. Additionally, we ablate \methodname{} with the following conditions: 3) \textbf{Deterministic Autoencoder (DAE)} is a deterministic version of \methodname{}, where $\CS{}$ is a single vector output of $\SE{}(\cdot)$ without any KL constraint. At inference, we sample the style code from a unit Gaussian distribution, $\CS{} \sim \mc{N}(0,I)$. 4) \textbf{Decoder Style Concatenation} removes the \ac{adain} layers and provides style information via a direct concatenation between $\CS{}$ and $\mc{F}^{\mc{P}}_{t}$ before the photo-realistic image generation module. To enable concatenation, we reshape $\CS{}$ into a multi-channel tensor, with a dimension equal to $\mc{F}^{\mc{P}}_{t}$.

% 5) \textbf{No Style Loss} removes the style consisitency loss presented in Eq. \ref{eq:style_loss}. $\mc{D}_{\mc{C}}$ is consistent for all baselines. \todo{May not want to put in No Style Loss, depending on how it looks}

\textbf{Evaluation Metrics:} We evaluate our model and baselines for both diversity and realism. We randomly select 100 poses from the validation set and generate 50 styles for each pose. We evaluate with the following metrics: 1) \textbf{Pairwise LPIPS distance:} We compute the Learned Perceptual Image Patch Similarity (LPIPS) \cite{zhang2018unreasonable} between generated samples corresponding to each pose, and take the mean distance over all poses. A higher LPIPS value defines a larger style diversity in the generated images. 2) \textbf{FID:} We compute the Fr\'echet Inception Distance (FID) \cite{heusel2017gans} between the generated samples and a subset of the training data. FID captures how similar the distribution of generated samples is to the ground truth data, with a lower score representing higher realism. 

% \todo{Could add the Face Recognition here? If it works in time}

\begin{table}[b!]
\caption{Baseline model comparison for novel human appearance synthesis. We use LPIPS \cite{zhang2018unreasonable} distance between generated appearances to measure diversity and FID \cite{heusel2017gans} against a subset of the training data to measure realism.}
\centering
\resizebox{0.9\linewidth}{!}{%
\begin{tabular}{@{}p{2.0cm}cc@{}}
\toprule
& \multicolumn{1}{c}{LPIPS ($\uparrow$)} & \multicolumn{1}{c}{FID ($\downarrow$)} \\ \midrule
\multicolumn{1}{r|}{VUNet \cite{esser2018variational}} & $6x10^{-4}$ & 157.3 \\
\multicolumn{1}{r|}{Pix2PixHD + Noise \cite{isola2017image}} & 0.001 & 117.34 \\ 
\multicolumn{1}{r|}{DAE} & 0.169 & 81.34 \\ 
\multicolumn{1}{r|}{Decoder Concat.} & $2x10^{-5}$ & 103.4 \\ \hline
% \multicolumn{1}{r|}{No Style Loss} & &  \\ \hline
\multicolumn{1}{r|}{\textbf{\methodname{} (Ours)}} & \textbf{0.243} & \textbf{49.48}  \\
\bottomrule
\end{tabular}%
}
\label{tab:baseline}
\end{table}

% & \multicolumn{1}{c}{Face Recog. ($\downarrow$)}

\subsection{\textbf{Quantitative Evaluation}}

\textbf{Baseline Comparisons:} We first evaluate \methodname{} for the synthesis of novel human appearances. Table~\ref{tab:baseline} shows that \methodname{} significantly outperforms all baselines in terms of both diversity and realism. VUnet \cite{esser2018variational} struggles with the large appearance diversity in the dataset, with the low LPIPS performance showing it fails to generate diverse appearances. Pix2PixHD+Noise \cite{isola2017image} generates more realistic images as indicated by the higher FID score, but the lack of control in the stochastic noise input during training results in a generation of only appearances seen during training.

Additionally, ablation results show the importance of the proposed \methodname{} model. A removal of the conditional variational autoencoder framework in the DAE setup results in a style distribution not centered around a prior. Even though realistic appearances can be generated via sampling, they are often of a consistent appearance lacking in diversity as shown in the weaker LPIPS score. The use of decoder concatenation instead of \ac{adain} layers is not powerful enough to encode style in the generated image, as shown by the low LPIPS performance.

\textbf{Realism Perceptual Study:} Our first perceptual study evaluates the realism of our novel human appearance synthesis. We show each participant 10 pairs of images, each containing a random shuffle of a real appearance and a sampled novel appearance. The participant is asked to select which appearance they believe is real, on a Likert scale from 1 to 5, where 1 is \textit{`left image is definitely real'}, 2 is \textit{`left image is probably real'}, 3 is \textit{`both left and right look similarly as real'} etc. We generate both sets of images using the photo-realistic generator of \methodname{}, either from an extracted (real) or sampled (novel) style code, to remove other factors that may affect the decision. We note similar perceptual evaluation has been used in \cite{isola2017image}. In total, 24 participants completed the study. Our novel appearances were given a score of 3.36 on average, a majority choice of \textit{`both left and right look similarly as real'}. Discretising the Likert scale down to a preference for each appearance, 23\% of participants believed our novel appearances were real (Likert choice 1 or 2) and 32\% thought they were equally as real as the original appearances (Likert choice 3). This highlights the realism of our novel human appearance synthesis for the \ac{slva} task.

% \begin{table}[b!]
% \caption{Realism Perceptual Study results}
% \centering
% \resizebox{0.85\linewidth}{!}{%
% \begin{tabular}{@{}p{0.0cm}cc@{}}
% \toprule
% & \multicolumn{2}{c}{Participant Selection} \\ 
% & \multicolumn{1}{c}{\textit{Real} (\%)} & \multicolumn{1}{c}{\textit{Fake} (\%)} \\ \midrule
% \multicolumn{1}{r|}{Real appearance} &  &  \\
% \multicolumn{1}{r|}{Generated appearance} &  &  \\
% \bottomrule
% \end{tabular}%
% }
% \label{tab:realism_perceptual}
% \end{table}

\begin{table}[b!]
\caption{Anonymity perceptual study results, showing the percentage of time participants chose a novel appearance to be most similar to a \textit{base appearance} ($\mc{I}^{\mc{S}1}$ or $\mc{I}^{\mc{S}2}$), a \textit{random appearance} or \textit{none}.}
\centering
\resizebox{0.91\linewidth}{!}{%
\begin{tabular}{@{}p{0.0cm}ccc@{}}
\toprule
& \multicolumn{3}{c}{Participant Selection} \\ 
& \multicolumn{1}{c}{\textit{Base}} & \multicolumn{1}{c}{\textit{Random}} & \multicolumn{1}{c}{\textit{None}} \\ \midrule
\multicolumn{1}{r|}{Novel Appearances} & 10.5\% & 57.5\% & 32.0\% \\
\bottomrule
\end{tabular}%
}
\label{tab:anonymity_perceptual}
\end{table}

\begin{table}[b!]
\caption{Downstream \ac{slt} results on the \ac{ph14t} dataset.}
\centering
\resizebox{0.88\linewidth}{!}{%
\begin{tabular}{@{}p{0.0cm}cc|cc@{}}
\toprule
     & \multicolumn{2}{c}{DEV SET}  & \multicolumn{2}{c}{TEST SET} \\ 
\multicolumn{1}{c|}{Data:}  & BLEU-4   & ROUGE  & BLEU-4  & ROUGE  \\ \midrule

\multicolumn{1}{r|}{Original} & 13.75 & 32.81 & 13.52 & 32.14 \\ 
\multicolumn{1}{r|}{Anonymised} & 13.58 & 32.12 & 13.09 & 30.24 \\ 

\bottomrule
\end{tabular}%
}
\label{tab:downstream_SLT}
\end{table}

\begin{figure*}[t!]
    \centering
    \includegraphics[width=0.9\linewidth]{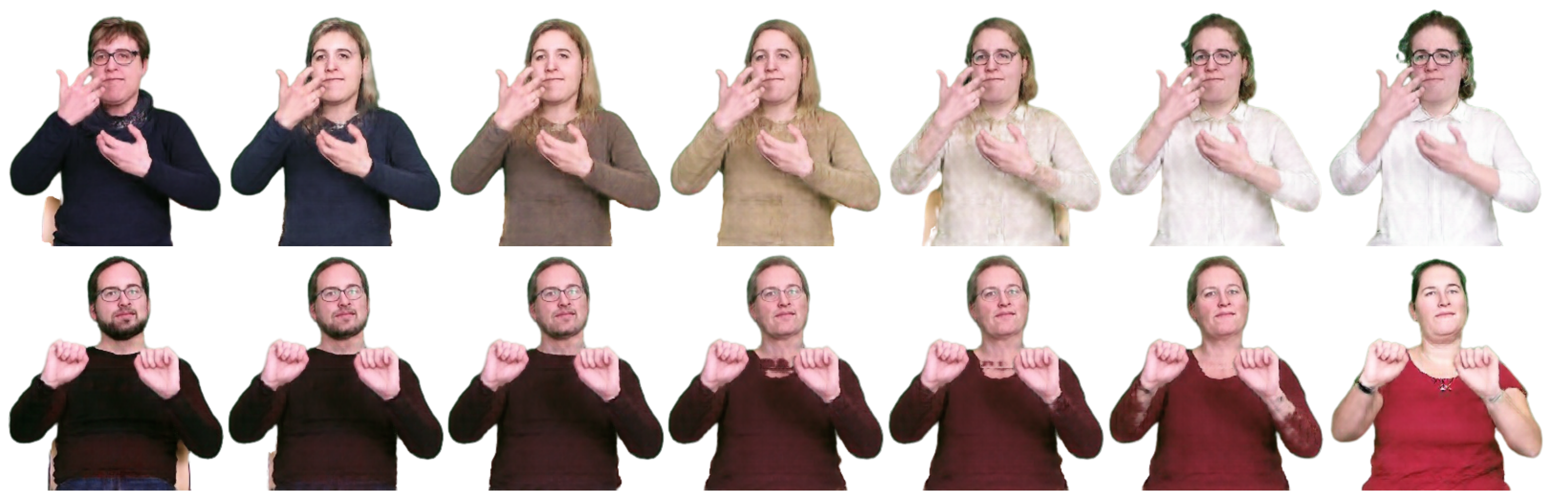}
    \caption{Style code traversal between two original appearances generates novel human appearances using \methodname{}}
    \label{fig:style_code_traversal}
\end{figure*}%

\begin{figure*}[b!]
    \centering
    \includegraphics[width=0.9\linewidth]{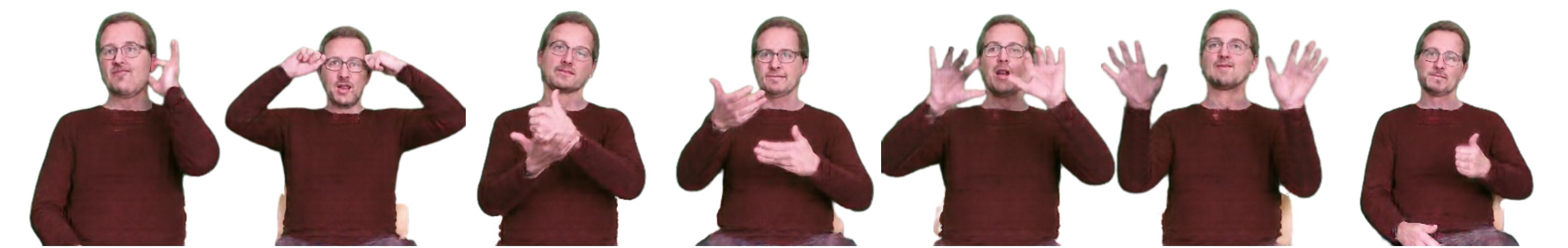}
    \caption{Style consistency over a pose sequence is achieved with our proposed \textit{style loss}}
    \label{fig:pose_sequence}
\end{figure*}%

\textbf{Anonymity Perceptual Study:} Determining whether a novel appearance resembles the original data is an inherently subjective task. Taking inspiration from super-recogniser tests \cite{dunn2020unsw}, we formalise an anonymity perceptual study for the \ac{slva} task that uses a photo lineup. We generate a novel appearance by blending between the style codes of two \textit{base appearances}, $\mc{I}^{\mc{S}1}$ and $\mc{I}^{\mc{S}2}$, thus generating an appearance with a mixed style, $\mc{Y}^{mix}$. Formally, we define the mixing result between style codes $\mc{C}^{\mc{S}1}$ and $\mc{C}^{\mc{S}2}$ as:
\begin{equation} \label{eq:style_mixing}
    \mc{C}^{mix} = \beta \mc{C}^{\mc{S}1} + (1 - \beta) \mc{C}^{\mc{S}2}
\end{equation}
where $\beta \in (0,1)$ and $\beta$ decreases from 1 to 0 in uniform steps. To evaluate the anonymity of this novel appearance, we show the participant a selection of 8 original appearances found in the training dataset. Two of these appearances are the \textit{base} style images used to generate the novel appearance, $\mc{I}^{\mc{S}1}$ and $\mc{I}^{\mc{S}2}$, and the other 6 are appearances randomly selected from the training dataset. We ask the participant to choose which original appearance looks most like the novel appearance, with an option for none of the appearances. 

We suggest that the novel appearance is anonymous if it bears a closer resemblance to a different original appearance than $\mc{I}^{\mc{S}1}$ or $\mc{I}^{\mc{S}2}$. Table \ref{tab:anonymity_perceptual} provides results of our anonymity perceptual study, showing that participants chose the correct \textit{base} appearance only 10.5\% of the time. 57.5\% of the time the participant chose a random original appearance, with a 32\% choice of none. This highlights that \methodname{} is able to generate convincingly novel human appearances that do not resemble the original appearances it was sampled from, suggesting strong anonymity performance for the \ac{slva} task.

\textbf{Downstream Translation Task:} To evaluate the retention of sign language content when performing video anonymisation, we perform a downstream \ac{slt} task using both the original and anonymised data. We use the challenging \ac{ph14t} dataset released by Camgoz \etal \cite{camgoz2018neural} and report BLEU and ROUGE scores. We implement the state-of-the-art \ac{slt} model \cite{camgoz2020sign} as our translation backbone and represent the photo-realistic videos using EfficientNet-B7 features \cite{tan2019efficientnet}.

Table \ref{tab:downstream_SLT} shows that sign language data anonymised by \methodname{} achieves comparable \ac{slt}  \hbox{BLEU-4} scores to the original data for both the development (13.75 vs. 13.58) and test (13.52 vs. 13.09) sets. This suggests that visual anonymisation using \methodname{} retains the original sign language content.

\subsection{\textbf{Qualitative Evaluation}}

\textbf{Baseline Comparisons:} Baseline comparisons are shown in Fig. \ref{fig:baselines}. It can be seen that \methodname{} generates realistic and diverse novel appearances more consistently than baselines. VUNet \cite{esser2018variational} struggles to replicate the high-quality and diverse nature of the SMILE dataset, resulting in a blended generation of blurry appearances. Pix2PixHD+Noise \cite{isola2017image} generates realistic samples, but experiences mode collapse with a replication of only appearances seen during training.

DAE generates high-quality outputs with little diversity. Due to its deterministic training, the learnt style manifold has an arbitrary unknown prior distribution. This makes the model difficult to sample from, resulting in a consistent appearance synthesis. Decoder style concatenation generates a consistent appearance regardless of the style code, showing that the configuration does not correctly encode style.

\textbf{Style Code Traversal:} Using a trained \methodname{} model, we can traverse along the style code manifold between two individuals, shown in Eq. \ref{eq:style_mixing} as a blend between style codes $\mc{C}^{\mc{S}1}$ to $\mc{C}^{\mc{S}2}$. Two style code traversal examples are shown in Fig. \ref{fig:style_code_traversal}. It can be seen that interpolating between known appearances provides a way to create realistic novel human appearances with a blended appearance.

\textbf{Style Consistency:} Due to the proposed \textit{style loss}, the sample of a novel appearance results in a consistent style over multiple frames and poses. Fig. \ref{fig:pose_sequence} shows an example of style consistency across a pose sequence.

% \section{Limitations} \label{sec:conclusion}

% \todo{Do we need this section?}

% \begin{itemize}
%     \item Bias towards white female signers in the dataset
%     \item Spurious interleaving of body parts or garments in the generated images (See fig XX)
% \end{itemize}

% \section{Online Sign Language Anonymisation} \label{sec:conc}

% \todo{Small discussion on anonymisation of online video content, to post anonymously/talk openly on twitter? Is this appropriate to say? Or could just add in some more qualitative figures instead?}

% In this section we discuss the potential application of \methodname{} to the visual anonymisation of online sign language data. Spoken language users are able to easily anonymise themselves online, due to the written form of language not revealing any personal information. In contrast, the Deaf are unable to share sign language content online without sharing their visual appearance and connected identifiable features. This may impact the openness of their speech, affecting the 

% We believe the application of \ac{slva} could increase the freedom of sign language data shared online, allowing important voices to be heard and opinions to be shared. Visual sign language anonymisation is akin to written text, where sign language content can be retained with a removal of any personal information. This may enable social media to serve the Deaf community in a more positive way without any privacy concerns.

\section{Conclusion} \label{sec:conc}

In this paper we formally introduced the task of \acf{slva}, an essential step towards addressing the privacy concerns risen whilst collecting large-scale sign language datasets. Furthermore, we proposed \methodname{}, a novel automatic method for the visual anonymisation of sign language data with a retention of the original sign language content. Using skeletal pose extracted from the source video, \methodname{} generates an anonymised photo-realistic video using a conditional variational autoencoder-based approach.

To enable the generation of novel signer appearances not seen during training, we learnt an approximate posterior style distribution over input signer appearances. At inference, we sampled from the style distribution to decode a photo-realistic anonymous signer appearance. In addition, we proposed a novel \textit{style loss} that ensures a style consistency in the generated anonymous video.

As future work, we aim to learn a structured latent space to ease customisation of our novel human appearance synthesis. In addition, we wish to tackle the anonymisation of person-specific signing styles that could be used to identify the original signer.

%------------------------------------------------------------------------
\section{Acknowledgements}
This project was supported by the EPSRC project ExTOL (EP/R03298X/1), the SNSF project SMILE2 (CRSII5\_193686) and the EU project EASIER (ICT-57-2020-101016982). This work reflects only the authors view and the Commission is not responsible for any use that may be made of the information it contains. We would also like to thank NVIDIA Corporation for their GPU grant.

% \newpage
{\small
\bibliographystyle{ieee}
\bibliography{bibliography}
}

\end{document}

%% file: acronyms.tex
\begin{acronym}[PHOENIX14\textbf{T} ]

\acrodefplural{rnn}[RNNs]{Recurrent Neural Networks}
\acrodefplural{cnn}[CNNs]{Convolutional Neural Networks}
\acrodefplural{hmm}[HMMs]{Hidden Markov Models}
\acrodefplural{gru}[GRUs]{Gated Recurrent Units}
\acrodefplural{crf}[CRFs]{Conditional Random Fields}
\acrodefplural{gan}[GANs]{Generative Adversarial Networks}
\acrodefplural{gpu}[GPUs]{Graphic Processing Units}

\acrodefplural{mdn}[MDNs]{Mixture Density Networks}

\acrodefplural{vae}[VAEs]{Variational Autoencoders}

\acro{adain}[AdaIN]{Adaptive Instance Normalisation}

\acro{bsl}[BSL]{British Sign Language}
\acro{bleu}[BLEU]{Bilingual Evaluation Understudy}
\acro{blstm}[BLSTM]{Bidirectional Long Short-Term Memory}
\acro{cnn}[CNN]{Convolutional Neural Network}
\acro{crf}[CRF]{Conditional Random Field}
\acro{cslr}[CSLR]{Continuous Sign Language Recognition}
\acro{ctc}[CTC]{Connectionist Temporal Classification}
\acro{dl}[DL]{Deep Learning}
\acro{dgs}[DGS]{German Sign Language - Deutsche Gebärdensprache}
\acro{dsgs}[DSGS]{Swiss German Sign Language - Deutschschweizer Geb\"ardensprache}
\acro{dtw}[DTW]{Dynamic Time Warping}
\acro{fc}[FC]{Fully Connected}
\acro{ff}[FF]{Feed Forward}
\acro{gan}[GAN]{Generative Adversarial Network}
\acro{gpu}[GPU]{Graphics Processing Unit}
\acro{gru}[GRU]{Gated Recurrent Unit}
\acro{gtpt}[G2PT]{Gloss to Pose Transformer}
\acro{hmm}[HMM]{Hidden Markov Model}
\acro{hpe}[HPE]{Hand Pose Enhancer}
\acro{isl}[ISL]{Irish Sign Language}
\acro{lstm}[LSTM]{Long Short-Term Memory}
\acro{mha}[MHA]{Multi-Headed Attention}
\acro{mtc}[MTC]{Monocular Total Capture}
\acro{mse}[MSE]{Mean Squared Error}
\acro{mdn}[MDN]{Mixture Density Network}
\acro{moe}[MoE]{Mixture-of-Expert}
\acro{nmt}[NMT]{Neural Machine Translation}
\acro{nlp}[NLP]{Natural Language Processing}
\acro{ph12}[PHOENIX12]{RWTH-PHOENIX-Weather-2012}
\acro{ph14}[PHOENIX14]{RWTH-PHOENIX-Weather-2014}
\acro{ph14t}[PHOENIX14\textbf{T}]{RWTH-PHOENIX-Weather-2014\textbf{T}}
\acro{pof}[POF]{Part Orientation Field}
\acro{paf}[PAF]{Part Affinity Field}
\acro{pttt}[P2TT]{Pose to Text Transformer}
\acro{relu}[RELU]{Rectified Linear Units}
\acro{rnn}[RNN]{Recurrent Neural Network}
\acro{rouge}[ROUGE]{Recall-Oriented Understudy for Gisting Evaluation}
\acro{sgd}[SGD]{Stochastic Gradient Descent}
\acro{sla}[SLA]{Sign Language Assessment}
\acro{slr}[SLR]{Sign Language Recognition}
\acro{slt}[SLT]{Sign Language Translation}
\acro{slp}[SLP]{Sign Language Production}
\acro{smt}[SMT]{Statistical Machine Translation}
\acro{slva}[SLVA]{Sign Language Video Anonymisation}

\acro{ttgt}[T2GT]{Text to Gloss Transformer}
\acro{ttpt}[T2PT]{Text to Pose Transformer}
\acro{ttp}[T2P]{Text to Pose}
\acro{ttgtp}[T2G2P]{Text to Gloss to Pose}

\acro{vae}[VAE]{Variational Autoencoder}

\acro{wer}[WER]{Word Error Rate}

\end{acronym}